\documentclass[10pt,conference]{IEEEtran}
\IEEEoverridecommandlockouts
\usepackage{cite}
\usepackage{amsmath,amssymb,amsfonts,amsthm}
\usepackage{algorithmic}
\usepackage{graphicx}
\usepackage{textcomp}
\usepackage{xcolor}
\usepackage{color}

\usepackage[linesnumbered,ruled,vlined]{algorithm2e}

\SetKwInOut{Parameter}{Parameter}

\usepackage{url}

\usepackage{subfigure}
\usepackage{csquotes}
\usepackage{enumitem}
\usepackage{romannum}
\usepackage{siunitx}
\usepackage{multirow}
\usepackage{float}
\usepackage{stfloats}
\usepackage[normalem]{ulem}
\useunder{\uline}{\ul}{}
\usepackage{svg}
\newcommand{\ratio}{\emph{\textrm{ratio}}}
\newcommand{\fr}{\alpha}
\newcommand{\offset}{\emph{\textrm{offset}}}

\usepackage[a4paper, total={184mm,239mm}]{geometry}

\def\BibTeX{{\rm B\kern-.05em{\sc i\kern-.025em b}\kern-.08em
    T\kern-.1667em\lower.7ex\hbox{E}\kern-.125emX}}

\begin{document}

\title{{
Timely Fusion of Surround Radar/Lidar for Object Detection in Autonomous Driving Systems}
% *\\
% \thanks{This work is partially supported by the Research Grants Council of Hong Kong (GRF 11208522, 15206221).}
}

% \author{{\em omitted for blind review}}

\author{\IEEEauthorblockN{Wenjing Xie}
\IEEEauthorblockA{
\textit{City University of Hong Kong}\\
wenjing.xie@my.cityu.edu.hk}
\and
\IEEEauthorblockN{Tao Hu}
\IEEEauthorblockA{\textit{City University of Hong Kong}\\
taohu9-c@my.cityu.edu.hk}
\and
\IEEEauthorblockN{Neiwen Ling}
\IEEEauthorblockA{\textit{Yale University}\\
neiwen.ling@yale.edu}
\and
\IEEEauthorblockN{Guoliang Xing}
\IEEEauthorblockA{\textit{The Chinese University of Hong Kong}\\
glxing@cuhk.edu.hk}
\and
\IEEEauthorblockN{Chun Jason Xue}
\IEEEauthorblockA{\textit{Mohamed bin Zayed University of Artificial Intelligence}\\
jason.xue@mbzuai.ac.ae}
\and
\IEEEauthorblockN{Nan Guan}
\IEEEauthorblockA{\textit{City University of Hong Kong}\\
nanguan@cityu.edu.hk}}

\maketitle

\begin{abstract}
Fusing Radar and Lidar sensor data can fully utilize their complementary advantages and provide more accurate reconstruction of the surrounding for autonomous driving systems.
Surround Radar/Lidar can provide \ang{360} view sampling with the minimal cost, which are promising sensing hardware solutions for autonomous  driving systems.
However, due to the intrinsic physical constraints, the rotating speed of surround Radar, and thus the frequency to generate Radar data frames, is much lower than surround Lidar. Existing Radar/Lidar fusion methods have to work at the low frequency of surround Radar, which cannot meet the high responsiveness requirement of autonomous driving systems.
This paper develops techniques to fuse surround Radar/Lidar with working frequency only limited by the faster surround Lidar instead of the slower surround Radar, based on widely-used object detection model called MVDNet. The basic idea of our approach is simple: we let MVDNet work with temporally unaligned data from Radar/Lidar, so that fusion can take place at any time when a new Lidar data frame arrives, instead of waiting for the slow Radar data frame. However, directly applying MVDNet to temporally unaligned Radar/Lidar data greatly degrades its object detection accuracy. The key information revealed in this paper is that we can achieve high output frequency with little accuracy loss by enhancing the training procedure to explore the temporal redundancy in MVDNet so that it can tolerate the temporal unalignment of input data. We explore several different ways of training enhancement and compare them quantitatively with experiments.
\end{abstract}

% \begin{IEEEkeywords}
%  Radar/Lidar fusion, object detection, autonomous driving 
% \end{IEEEkeywords}

\section{Introduction}
Accurate perception of the surroundings 
is a fundamental requirement for 
autonomous driving systems (ADS). Among the various sensors used in ADS, Lidar and Radar have emerged as highly desirable options for perception tasks \cite{zhang2020sensing}. 
% While Lidar sensors can generate fine-grained point clouds with rich information in good weather conditions, they fail in adverse weather (e.g., fog, snow) \cite{qian2021robust}. By contrast, Radar sensors are less affected by adverse weather conditions, but they exhibit lower precision compared to Lidar \cite{campbell2018sensor}.
% To overcome the limitations of individual sensors and enhance the accuracy of object detection, fusing Lidar and Radar data has shown promising results in providing accurate and robust perception capability for ADS \cite{kocic2018sensors,wang2019multi}.
While Lidar provides detailed point clouds in favorable weather conditions, it struggles in adverse weather (e.g., fog, snow), whereas Radar is less affected by adverse conditions but offers lower precision \cite{qian2021robust}. To overcome these limitations, fusing Lidar and Radar data has shown promise in enhancing object detection accuracy.

To obtain a comprehensive view of the surrounding environment, it is necessary for both Lidar and Radar sensors to cover the entire \ang{360} angle. 
One traditional approach is to mount multiple sensors with different orientations, each covering a specific angle. The data from these sensors can then be combined to construct a \ang{360} view. However, this approach comes with the drawback of high overall costs due to the requirement of multiple sensors and their associated hardware.
Alternatively, a more cost-effective solution is to utilize surround Radar/Lidar systems to capture the full \ang{360} view.
The temporal resolution of surround Radar/Lidar, i.e., the frequency at which they produce full-view data frames, is limited by the rotating speed. Radar, which operates using millimeter-waves, travels much slower than Lidar, which uses light, resulting in lower rotating speeds for Radar under similar operating conditions.
For example, a popular surround Lidar sensor, Velodyne HDL-32E, rotates at a frequency of $20$Hz ($20$ cycles per second) \cite{HDL32E}, while the state-of-the-art surround Radar sensor, NavTech CTS350-X, rotates at a frequency of $4$Hz ($4$ cycles per second) \cite{CTS350X}.
Consequently, it is necessary to address the inconsistent input frequencies between surround Radar and Lidar when fusing their data. One straightforward solution is to down-sample the Lidar data to match the Radar frequency. Recent research has also explored combining consecutive Lidar data frames to produce artificial Lidar frames that align with the low frequency of Radar \cite{qian2021robust,li2022modality}. In both cases, fusion is performed at the frequency dictated by the Radar rotation. 
For instance, with a surround Lidar rotating at $20$Hz and a surround Radar rotating at $4$Hz, fusion is performed and thus generates full-view reconstruction $4$ times per second. Typically, ADS requires to capture events and take proper reaction in a short time, e.g., $100$ms. Therefore, the low frequency of surround Radar/Lidar fusion makes it unsuitable for ADS from the real-time performance perspective. 

In this paper, we propose techniques to address the aforementioned problem of low fusion frequency in Radar/Lidar fusion. Our approach aims to increase the fusion frequency beyond the limitations imposed by the low rotation speed of surround Radar. 
Our work is based on MVDNet, a widely-used Radar/Lidar data fusion deep neural network model in object detection. The core idea of our approach is to increase the fusion frequency by fusion the latest available Radar data frame with a new Lidar frame. 
However, this introduces temporal inconsistency between the Radar and Lidar sensor data, which can significantly degrade detection accuracy. To address this, we explore different methods to enhance the training procedure and evaluate their effectiveness through quantitative experiments.

\section{Related Work}

Many studies have focused on fusing different sensor modalities, such as cameras, Lidars, and Radars, to improve 3D perception in autonomous driving. MVDNet \cite{qian2021robust} employs an attention mechanism to fuse Radar and Lidar data for vehicle detection in foggy weather and achieves state-of-the-art results on the Oxford Radar RobotCar (ORR) dataset \cite{barnes2020oxford}. Other works, such as ST-MVDNet \cite{li2022modality}, propose methods to handle missing sensor data in multi-modal vehicle detection. Real-time Radar/Lidar fusion methods for road-object detection and tracking in ADS have been proposed by Farag et al. \cite{farag2021real}. LiRaNet \cite{shah2020liranet} and EZFfusion \cite{li2022mathsf} focus on end-to-end trajectory prediction and multi-modal 3D object detection and tracking, respectively. DEF \cite{bijelic2020seeing} and FUTR3D \cite{chen2022futr3d} introduce fusion frameworks applicable to various sensor configurations. Bi-LRFusion \cite{wang2023bi} enhances 3D detection for dynamic objects through bi-directional LiDAR-Radar fusion. LiRaFusion \cite{song2024lirafusion} proposes early and middle fusion modules for joint voxel feature encoding and adaptive feature map fusion.
However, existing Radar and Lidar fusion models assume perfect data synchronization. Down-sampling the faster Lidar data to match the Radar frequency is a common approach \cite{qian2021robust,li2022modality,kim2020low,shah2020liranet,farag2021real,li2022mathsf,bijelic2020seeing,chen2022futr3d}. Some researchers have explored real-time infrastructure-vehicle cooperative approaches for timely fusion and perception \cite{prakash2021multi,cui2022coopernaut,he2021vi}. In this paper, we propose techniques to fuse temporally unaligned Radar/Lidar data and achieve a high output frequency with minimal accuracy loss.
\section{Preliminary}
\label{Smvdnet}
\begin{figure}[htbp]
\centerline{\includegraphics[scale=0.5]{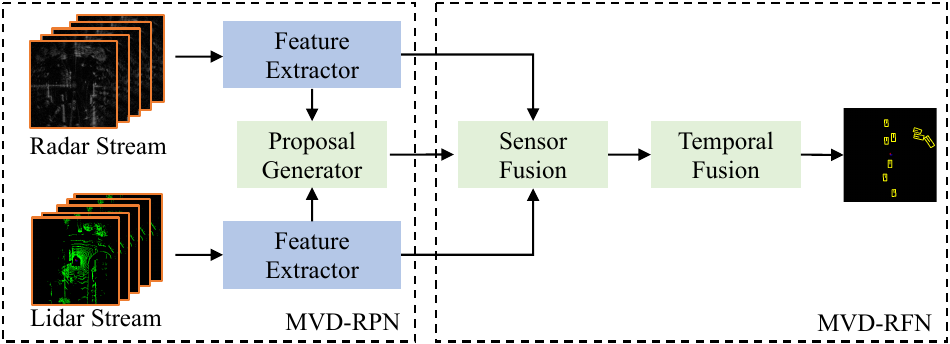}}
\vspace{-0.08in}
\caption{The architecture of MVDNet\cite{qian2021robust}.}
\label{mvdnet}
\vspace{-0.1in}
\end{figure}

\begin{figure}
\centerline{\includegraphics[scale=0.48]{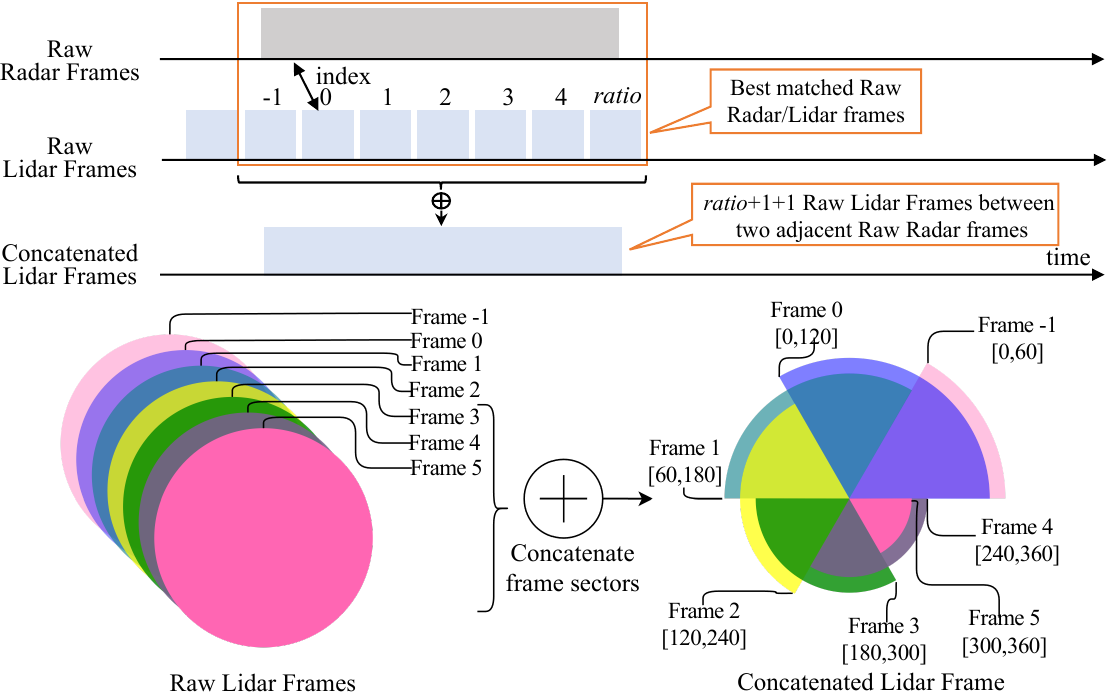}}
\vspace{-0.1in}
\caption{The Radar/Lidar frames alignment policy in MVDNet.}
\label{sector}
\vspace{-0.2in}
\end{figure}

We build our work based on MVDNet \cite{qian2021robust}, first object detection network that fuses Lidar and high-resolution \ang{360} radar signals.
% the state-of-the-art Radar/Lidar fusion network.
MVDNet fuses Radar intensity maps with Lidar point clouds, which harnesses their complementary capabilities. 
The architecture of MVDNet is shown in Fig.~\ref{mvdnet}. 
MVDNet consists of two stages. 
The region proposal network (MVD-RPN) extracts feature maps from Lidar and Radar inputs and generates proposals. The region fusion network (MVD-RFN) pools and fuses region-wise features of the two sensors' data frames and outputs oriented bounding boxes of detected vehicles.
\vspace{-0.1in}
%figure: MVDNet architecture
MVDNet assumes that the input Radar and Lidar frames have same timestamps. 
However, in the training data of MVDNet, 
the frequency of raw Radar frames, denoted by $F_r$, is different from the frequency of raw Lidar frames, denoted by $F_l$.
MVDNet solves this problem by combining several consecutive raw Lidar frames into an artificial \emph{concatenated Lidar frame},
which is generated with the same frequency as Radar data frames, as shown in Fig.~\ref{sector}.
In this way, the fusion frequency of MVDNet equals $F_r$.
We define 
\[
\ratio = \left \lfloor \frac{F_l}{F_r} \right \rfloor
\]
and \emph{ratio}+1+1 raw Lidar frames are aggregated to produce 
one concatenated Lidar data frame.
More precisely, a subset of cloud points that covers \ang{120} view in each Lidar frame is selected, 
and these subsets are combined into a concatenated Lidar frame as shown in Fig.~\ref{sector}.
Each concatenated Lidar frame and corresponding Radar frame are paired and sent to MVDNet.
To improve accuracy, when fusing the latest paired Radar/Lidar frames, MVDNet uses historical frames. The number of historical frames is denoted by \emph{num\_history}.
In MVDNet, \emph{num\_history}=4 by default. Decreasing \emph{num\_history} can reduce the computation workload of MVDNet inference, but at the cost of accuracy loss, as discussed in \ref{ss:history} and \ref{ss:frequency}.

\section{The Proposed Method}

\subsection{Dataset Preparation}
The original ORR data are collected by a vehicle
equipped with a NavTech CTS350-X Radar \cite{CTS350X} at the roof
center, co-located with two Velodyne HDL-32E \cite{HDL32E} Lidars
whose point clouds are combined. 
MVDNet \cite{qian2021robust} manually generates the ground-truth labels based on the ORR Lidar point clouds. They create 3D bounding boxes for vehicles in one out of every 20 frames (i.e., 1s) using the Scalable annotation tool. Labels of the remaining 19 frames are interpolated using the visual odometry data provided in ORR and manually adjusted to align with the corresponding vehicles.
To demonstrate the effectiveness of our proposed timely fusion methods, we use the same processed dataset as MVDNet.

\subsection{Increasing the Radar/Lidar Fusion Frequency}
\label{ss:frequency}
% figure: examples for fusion 
\begin{figure*}
\centering
	\subfigure[The first example, $\alpha$=1,$F_f$=$F_l$.]{
		\begin{minipage}{18cm} 
		\label{example1}
        \centerline{\includegraphics[scale=0.5]{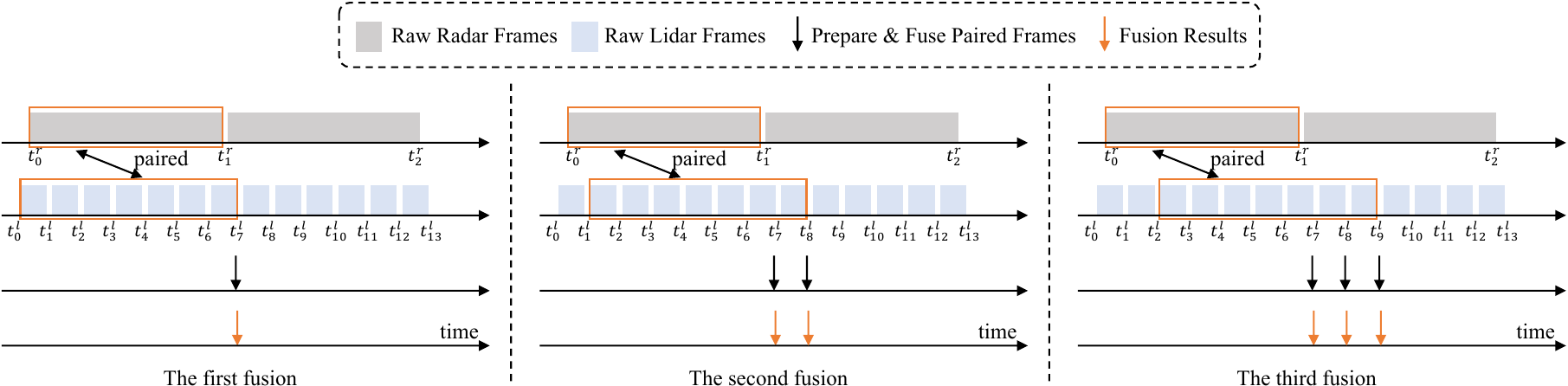}}
        \vspace{0.1in}
		\end{minipage}
	}
 
\vspace{-0.1in}	

	\subfigure[The second example, $\alpha$=3,$F_f$=$F_l$/3.]{
		\begin{minipage}{18cm}%[b]%{0.2\textwidth}
		\label{example2}
		\centerline{\includegraphics[scale=0.5]{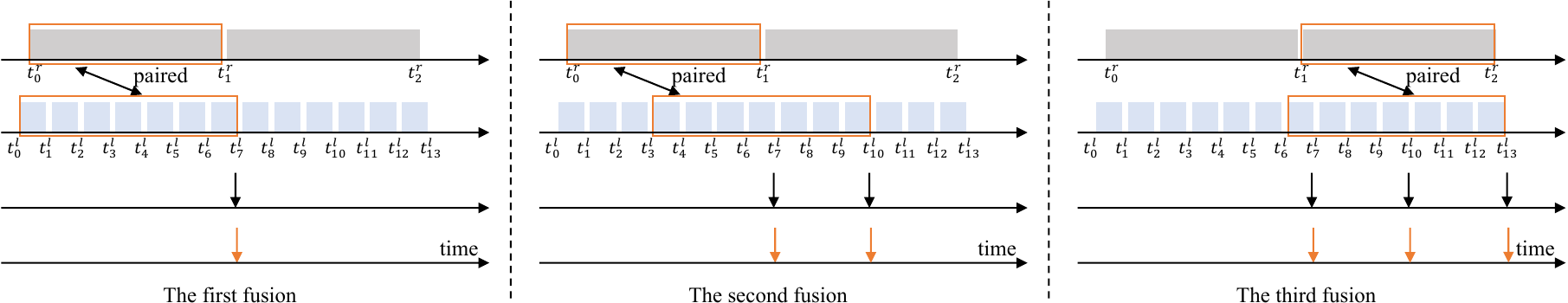}}
		\vspace{0.1in}
		\end{minipage}
	}

\vspace{-0.08in}
\caption{Our proposed fusion method.}
\label{example}
\vspace{-0.2in}
\end{figure*}

%para1
% Our method enables immediate fusion upon the arrival of a raw Lidar frame, allowing for a higher fusion frequency compared to waiting for slow Radar frames.
Our method enables fusion to be executed immediately upon the arrival of a raw Lidar frame, instead of waiting for the slow Radar frames, so fusion can be triggered with a higher frequency. 
The fusion frequency of our method, denoted as $F_f$, is different from existing methods \cite{qian2021robust,li2022modality,kim2020low,shah2020liranet,farag2021real,li2022mathsf,bijelic2020seeing,chen2022futr3d,prakash2021multi,cui2022coopernaut,he2021vi}, where the $F_f$ is equal to the Radar frequency $F_r$. In our method, the fusion frequency  $F_f$ can be increased within the range:
\begin{equation}
\label{frequency}
F_r \leq F_f \leq F_l
\end{equation}
Each arrival of a new raw Lidar frame provides an opportunity for fusion. In practice, fusion is feasible whenever the fusion period ($1/F_f$) is an integral number multiple of the Lidar frame period ($1/F_l$). This establishes a direct relationship between the fusion frequency and the Lidar frequency, represented by the parameter $\alpha$:
\begin{equation}
\label{alpha}
\alpha = \frac{F_l}{F_f}
\end{equation}
Where $\alpha$ should satisfy the condition:
\begin{equation}
\label{numFusion}
\fr \in \{1,2,3,...,ratio\}
\end{equation}
If $\alpha > \ratio$, the fusion of Radar and Lidar frames becomes meaningless due to the absence of temporal overlap. 
The fusion frequency in our method is limited only by the faster Lidar instead of the slower Radar.

Fig.~\ref{example} shows two examples to demonstrate our fusion method. When a new raw Lidar frame arrives and the matching raw Radar frame for corresponding interval has not yet arrived, our method utilizes the latest available raw Radar frame for fusion.
%example 1
In Fig. \ref{example1}, we aim to achieve a high fusion frequency $F_f=F_l$. According to Equation (\ref{alpha}), we need to set $\alpha$=1, indicating performing timely fusion when each raw Lidar frame arrives.
% the fusion is carried out immediately upon the 
% In Fig.~\ref{example1}, we assume that we expect to achieve a high fusion frequency $F_f=F_l$. According to Equation (\ref{alpha}), we need to set $\alpha$=1, which means to perform timely fusion when each new raw Lidar frame arrives. 
Consider that the first fusion event occurs at time $t_7^l$, when the second and third new raw Lidar frames arrive at time  $t_8^l$ and $t_9^l$ respectively, the Radar frame which is perfectly synchronized in time has not yet arrived and is expected to arrive at time $t_r^2$. Under these situations, fusion is conducted between the latest concatenated Lidar frames (the orange box) and the latest available raw Radar frame (the orange box) generated at $t_1^r$ respectively.
% In Fig.~\ref{example1}, the latest raw Radar frame has not been generated at $t_8^l$ and $t_9^l$. We perform the fusion between the latest concatenated Lidar frames (the orange box) and the latest available raw Radar frame (the orange box) generated at $t_1^r$ respectively.
Fig. \ref{example2} shows another example with fusion frequency $F_f=F_l/3$. The fusion should be performed every three raw Lidar arrives. 
Our method not only increases the fusion frequency but also allows us to adjust the frequency according to the actual situation. In practice, the fusion results do not always have to be generated at the highest frequency. Based on the current vehicle speed and overall workload of the vehicle, the ADS is expected to provide a fusion frequency that can be adjusted.

\subsection{Training Enhancement to Maintain Accuracy}
%table : directly apply mvdnet model on offset data
\begin{table*}[htbp]
\centering
\caption{Overall Performance of Directly Applying MVDNet on Radar/Lidar Frames: \protect\\AP of Oriented Bounding Boxes in Bird’s Eye View (BEV).}
\label{TdirectMVDNet}
\scalebox{0.82}{
\renewcommand{\arraystretch}{1.2} 
\begin{tabular}{|cc|ccc|ccc|ccc|ccc|ccc|ccc|}
\hline
\multicolumn{1}{|c|}{}                        & \emph{offset} & \multicolumn{3}{c|}{0}                                                                                                                                          & \multicolumn{3}{c|}{1}                                          & \multicolumn{3}{c|}{2}                                          & \multicolumn{3}{c|}{3}                                          & \multicolumn{3}{c|}{4}                                         & \multicolumn{3}{c|}{5}                                          \\ \cline{2-20} 
\multicolumn{1}{|c|}{\multirow{-2}{*}{Model}} & IoU         & \multicolumn{1}{c|}{0.5}                                   & \multicolumn{1}{c|}{0.65}                                  & 0.8                                   & \multicolumn{1}{c|}{0.5}   & \multicolumn{1}{c|}{0.65}  & 0.8   & \multicolumn{1}{c|}{0.5}   & \multicolumn{1}{c|}{0.65}  & 0.8   & \multicolumn{1}{c|}{0.5}   & \multicolumn{1}{c|}{0.65}  & 0.8   & \multicolumn{1}{c|}{0.5}  & \multicolumn{1}{c|}{0.65}  & 0.8   & \multicolumn{1}{c|}{0.5}   & \multicolumn{1}{c|}{0.65}  & 0.8   \\ \hline
\multicolumn{2}{|c|}{MVDNet}                                & \multicolumn{1}{c|}{{\color[HTML]{333333} \textbf{0.897}}} & \multicolumn{1}{c|}{{\color[HTML]{333333} \textbf{0.877}}} & {\color[HTML]{333333} \textbf{0.735}} & \multicolumn{1}{c|}{0.652} & \multicolumn{1}{c|}{0.619} & 0.486 & \multicolumn{1}{c|}{0.616} & \multicolumn{1}{c|}{0.587} & 0.462 & \multicolumn{1}{c|}{0.583} & \multicolumn{1}{c|}{0.559} & 0.441 & \multicolumn{1}{c|}{0.51} & \multicolumn{1}{c|}{0.498} & 0.411 & \multicolumn{1}{c|}{0.521} & \multicolumn{1}{c|}{0.508} & 0.407 \\ \hline
\end{tabular}
}

\vspace{0.2in}
% Table II
\caption{Overall Performance of Our Fusion Method on Radar/Lidar frames: AP of Oriented Bounding Boxes in BEV.}
% Bold numbers represent the best score among all models on the unaligned Radar/Lidar data with a specific time \emph{offset}.
\label{TourMVDNet}
\scalebox{0.82}{
\renewcommand{\arraystretch}{1.2} 
\begin{tabular}{|cc|ccc|ccc|ccc|ccc|ccc|}
\hline
\multicolumn{1}{|c|}{\multirow{2}{*}{Model}} & \textit{offset} & \multicolumn{3}{c|}{1}                                                                     & \multicolumn{3}{c|}{2}                                                                     & \multicolumn{3}{c|}{3}                                                                    & \multicolumn{3}{c|}{4}                                                                    & \multicolumn{3}{c|}{5}                                                                     \\ \cline{2-17} 
\multicolumn{1}{|c|}{}                       & IoU             & \multicolumn{1}{c|}{0.5}            & \multicolumn{1}{c|}{0.65}           & 0.8            & \multicolumn{1}{c|}{0.5}            & \multicolumn{1}{c|}{0.65}           & 0.8            & \multicolumn{1}{c|}{0.5}           & \multicolumn{1}{c|}{0.65}           & 0.8            & \multicolumn{1}{c|}{0.5}            & \multicolumn{1}{c|}{0.65}          & 0.8            & \multicolumn{1}{c|}{0.5}            & \multicolumn{1}{c|}{0.65}           & 0.8            \\ \hline
% \multicolumn{2}{|c|}{MVDNet}                                & \multicolumn{1}{c|}{0.652} & \multicolumn{1}{c|}{0.619} & 0.486 & \multicolumn{1}{c|}{0.616}          & \multicolumn{1}{c|}{0.587}          & 0.462          & \multicolumn{1}{c|}{0.583}         & \multicolumn{1}{c|}{0.559}          & 0.441          & \multicolumn{1}{c|}{0.51}          & \multicolumn{1}{c|}{0.498}         & 0.411          & \multicolumn{1}{c|}{0.521}          & \multicolumn{1}{c|}{0.508}          & 0.407          \\ \hline
\multicolumn{2}{|c|}{MVDNet\_1}                                & \multicolumn{1}{c|}{\textbf{0.898}} & \multicolumn{1}{c|}{\textbf{0.869}} & \textbf{0.722} & \multicolumn{1}{c|}{0.867}          & \multicolumn{1}{c|}{0.841}          & 0.668          & \multicolumn{1}{c|}{0.847}         & \multicolumn{1}{c|}{0.827}          & 0.652          & \multicolumn{1}{c|}{0.828}          & \multicolumn{1}{c|}{0.809}         & 0.643          & \multicolumn{1}{c|}{0.826}          & \multicolumn{1}{c|}{0.801}          & 0.644          \\ \hline
\multicolumn{2}{|c|}{MVDNet\_2}                                & \multicolumn{1}{c|}{0.87}           & \multicolumn{1}{c|}{0.85}           & 0.67           & \multicolumn{1}{c|}{\textbf{0.899}} & \multicolumn{1}{c|}{\textbf{0.877}} & \textbf{0.724} & \multicolumn{1}{c|}{0.889}         & \multicolumn{1}{c|}{0.863}          & 0.684          & \multicolumn{1}{c|}{0.876}          & \multicolumn{1}{c|}{0.848}         & 0.705          & \multicolumn{1}{c|}{0.868}          & \multicolumn{1}{c|}{0.847}          & 0.702          \\ \hline
\multicolumn{2}{|c|}{MVDNet\_3}                                & \multicolumn{1}{c|}{0.867}          & \multicolumn{1}{c|}{0.846}          & 0.699          & \multicolumn{1}{c|}{0.868}          & \multicolumn{1}{c|}{0.847}          & 0.701          & \multicolumn{1}{c|}{\textbf{0.89}} & \multicolumn{1}{c|}{\textbf{0.867}} & \textbf{0.722} & \multicolumn{1}{c|}{0.879}          & \multicolumn{1}{c|}{0.858}         & 0.709          & \multicolumn{1}{c|}{0.878}          & \multicolumn{1}{c|}{0.849}          & 0.7            \\ \hline
\multicolumn{2}{|c|}{MVDNet\_4}                                & \multicolumn{1}{c|}{0.877}          & \multicolumn{1}{c|}{0.855}          & 0.686          & \multicolumn{1}{c|}{0.881}          & \multicolumn{1}{c|}{0.853}          & 0.683          & \multicolumn{1}{c|}{0.874}         & \multicolumn{1}{c|}{0.853}          & 0.688          & \multicolumn{1}{c|}{\textbf{0.888}} & \multicolumn{1}{c|}{\textbf{0.86}} & \textbf{0.709} & \multicolumn{1}{c|}{0.875}          & \multicolumn{1}{c|}{0.854}          & 0.688          \\ \hline
\multicolumn{2}{|c|}{MVDNet\_5}                                & \multicolumn{1}{c|}{0.855}          & \multicolumn{1}{c|}{0.833}          & 0.679          & \multicolumn{1}{c|}{0.862}          & \multicolumn{1}{c|}{0.834}          & 0.687          & \multicolumn{1}{c|}{0.865}         & \multicolumn{1}{c|}{0.843}          & 0.689          & \multicolumn{1}{c|}{0.884}          & \multicolumn{1}{c|}{0.864}         & 0.704          & \multicolumn{1}{c|}{\textbf{0.885}} & \multicolumn{1}{c|}{\textbf{0.865}} & \textbf{0.707} \\ \hline
\end{tabular}
}
\vspace{-0.1in}
\end{table*}

\begin{figure}[htbp]
\vspace{-0.1in}
\centerline{\includegraphics[scale=0.5]{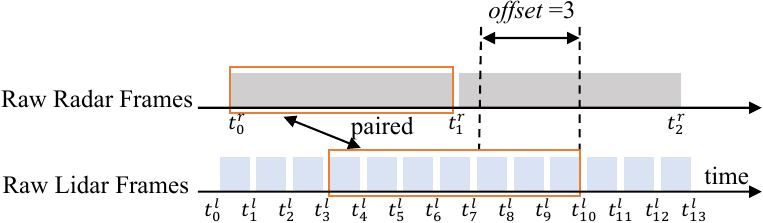}}
\vspace{-0.08in}
\caption{Offset between temporally unaligned frames.}
\label{offset}
\vspace{-0.2in}
\end{figure}

However, when directly applying the mentioned timely fusion method, there can be a temporal misalignment between the input Radar/Lidar frames to the MVDNet. This misalignment often leads to a significant decrease in object detection accuracy. 
In fact, each pair of unaligned Radar/Lidar frames exhibits a time offset, as illustrated in Fig. \ref{offset}. We use the term \emph{offset} to represent the number of complete raw Lidar frames that arrive ahead of the Radar in each paired Radar/Lidar frame set. The \emph{offst} must satisfy the condition:
\begin{equation}
\label{numOffset}
\offset \in \{0,1,2,...,\ratio\}
\end{equation}
A larger \emph{offset} value indicates a longer unaligned time interval between the paired Radar/Lidar frames. Consequently, if the \emph{offset} exceeds the \ratio, the fusion between the paired frames becomes meaningless because they do not overlap in time.

Following the MVDNet paper, we evaluate the original MVDNet model on unaligned Radar/Lidar frames using the ORR dataset ($F_l=20$Hz, $F_r=4$Hz, \emph{ratio}=5). To ensure a fair comparison, we utilize average precision (AP) in COCO evaluation \cite{lin2014microsoft} with Intersection-over-Union (IoU) thresholds of $0.5$, $0.65$, and $0.8$. The accuracy results for various \emph{offset} settings are shown in Table \ref{TdirectMVDNet}. Our observations are as follows:
\begin{itemize}
    \item The accuracy of the MVDnet model significantly decreases when applied to unaligned frames (\emph{offset} = $\{1,2,3,4,5\}$) compared to aligned frames (\emph{offset}=0) across different IoU thresholds. This indicates that using the original MVDNet model on unaligned Radar/Lidar frames significantly degrades object detection accuracy.
    \item Specifically, when considering a specific IoU value, such as IoU=$0.8$, we observe a further decrease in object detection accuracy as the \emph{offset} increases. This highlights the critical impact of the \emph{offset} on the accuracy of MVDNet.
\end{itemize}

%figure, describe num_offset
\vspace{-0.1in}

%para4
To address the issue of accuracy loss issue mentioned above, we leverage an important characteristic of MVDNet that incorporates both the current pair and a specific number of historical pairs (\emph{num\_history}=4) during fusion. This results in a total of 1+\emph{num\_history} paired Radar/Lidar frames being used as input for each fusion. 
It is important to note that there exists redundant information among consecutive Radar/Lidar frames. We exploit the redundant information to make MVDNet still work well with temporally unaligned Radar/Lidar frames. 
Furthermore, we discover a straightforward yet highly effective training enhancement method that enables MVDNet to maintain accuracy during the fusion of temporally unaligned Radar/Lidar frames. This method involves synthesizing paired Radar/Lidar frames with corresponding \emph{offset} values and utilizing them to train the MVDNet from scratch. By incorporating data with different \emph{offset} values during training, we aim to equip the MVDNet model with the ability to detect objects from temporally unaligned Radar/Lidar frames.

Our methods' results, obtained under the same experimental setting, are summarized in Table \ref{TourMVDNet}. MVDNet\_{\emph{offset}} refers to the MVDNet model trained on Radar/Lidar frames with a specific \emph{offset}, where \emph{offset}$\in\{1,2,3,4,5\}$. Our observations are as follows:
\begin{itemize}
  
    \item In table \ref{TourMVDNet}, for unaligned Radar/Lidar frames with a specific \emph{offset}=\emph{i}, the MVDNet\_\emph{i} model achieves better accuracy performance than other MVDNet\_\emph{j}, where $i, j \in \{1,2,3,4,5\}$ and $i != j$.
    \item Comparing Table \ref{TourMVDNet} and Table \ref{TdirectMVDNet}, the accuracy of our enhancement training method on unaligned Radar/Lidar frames with each specific \emph{offset} ($\emph{offset} \in \{1,2,3,4,5\}$) is almost equal to the original MVDNet model on aligned Radar/Lidar frames (\emph{offset}=0).
\end{itemize}
% These results demonstrate that, on the premise of increasing the fusion frequency, our proposed fusion method maintains accuracy even when the Radar/Lidar frames are time unaligned.

The results demonstrate that our proposed fusion method maintains accuracy even when the Radar/Lidar frames are temporally unaligned, while also increasing the fusion frequency.
% proposed fusion method maintains accuracy even when the Radar/Lidar frames are temporally unaligned, while simultaneously increasing the fusion frequency. 

\subsection{Exploiting Different Enhancement Strategies}
In order to enable MVDNet model to deal with Radar/Lidar frames with various \emph{offset}, we exploit two different enhancement strategies.

%separately training
\textbf{Separate training strategy:} 
For temporally unaligned Radar/Lidar frames with different \emph{offset}, the first strategy involves training a dedicated MVDNet model for each specific \emph{offset} frame. 
During training, we first train a shared MVDNet model using paired Radar/Lidar frames with \emph{offset}=$\lceil \ratio/2 \rceil$. The parameter weights of its MVD-RPN are then shared among other MVDNet models that handle different  \emph{offset} frames.
Next, the MVD-RFN of each MVDNet model is fine-tuned using the corresponding \emph{offset} frames. This process results in multiple MVDNet models, with each model designed to handle Radar/Lidar frames with a specific \emph{offset}.
The evaluation results on the ORR dataset are summarized in Table \ref{trainSeperate}. MVDNet\_separated\_\emph{offset} refers to an MVDNet model fine-tuned on Radar/Lidar frames with a specific \emph{offset}, where \emph{offset}$\in$\{0,1,2,...,5\}. We observe that for Radar/Lidar frames with a specific \emph{offset}, each MVDNet\_separated\_\emph{offset} model, fine-tuned on the corresponding \emph{offset} frames, achieves the best accuracy performance compared to other models.

% freeze MVDNet
\begin{table*}[htbp]
\centering

\caption{Overall performance of separately trained MVDNet models : AP of oriented bounding boxes in BEV.}
%Bold numbers represent the best score among all the models on Radar/Lidar frames with special time offset.}
\label{trainSeperate}
\scalebox{0.82}{
\renewcommand{\arraystretch}{1.2} 
\begin{tabular}{|cc|ccc|ccc|ccc|ccc|ccc|ccc|}
\hline
\multicolumn{1}{|c|}{\multirow{2}{*}{Model}} & \emph{offset} & \multicolumn{3}{c|}{0}                                                                     & \multicolumn{3}{c|}{1}                                                                     & \multicolumn{3}{c|}{2}                                                                     & \multicolumn{3}{c|}{3}                                                                     & \multicolumn{3}{c|}{4}                                                                     & \multicolumn{3}{c|}{5}                                                                     \\ \cline{2-20} 
\multicolumn{1}{|c|}{}                       & IoU         & \multicolumn{1}{c|}{0.5}            & \multicolumn{1}{c|}{0.65}           & 0.8            & \multicolumn{1}{c|}{0.5}            & \multicolumn{1}{c|}{0.65}           & 0.8            & \multicolumn{1}{c|}{0.5}            & \multicolumn{1}{c|}{0.65}           & 0.8            & \multicolumn{1}{c|}{0.5}            & \multicolumn{1}{c|}{0.65}           & 0.8            & \multicolumn{1}{c|}{0.5}            & \multicolumn{1}{c|}{0.65}           & 0.8            & \multicolumn{1}{c|}{0.5}            & \multicolumn{1}{c|}{0.65}           & 0.8            \\ \hline
\multicolumn{2}{|c|}{MVDNet\_separated\_0}                            & \multicolumn{1}{c|}{\textbf{0.899}} & \multicolumn{1}{c|}{\textbf{0.878}} & \textbf{0.736} & \multicolumn{1}{c|}{0.848}          & \multicolumn{1}{c|}{0.826}          & 0.679          & \multicolumn{1}{c|}{0.838}          & \multicolumn{1}{c|}{0.817}          & 0.675          & \multicolumn{1}{c|}{0.83}           & \multicolumn{1}{c|}{0.809}          & 0.68           & \multicolumn{1}{c|}{0.802}          & \multicolumn{1}{c|}{0.79}           & 0.671          & \multicolumn{1}{c|}{0.803}          & \multicolumn{1}{c|}{0.792}          & 0.663          \\ \hline
\multicolumn{2}{|c|}{MVDNet\_separated\_1}                            & \multicolumn{1}{c|}{0.887}          & \multicolumn{1}{c|}{0.866}          & 0.719          & \multicolumn{1}{c|}{\textbf{0.887}} & \multicolumn{1}{c|}{\textbf{0.866}} & \textbf{0.722} & \multicolumn{1}{c|}{0.884}          & \multicolumn{1}{c|}{0.864}          & 0.722          & \multicolumn{1}{c|}{0.884}          & \multicolumn{1}{c|}{0.856}          & 0.722          & \multicolumn{1}{c|}{0.876}          & \multicolumn{1}{c|}{0.855}          & 0.711          & \multicolumn{1}{c|}{0.874}          & \multicolumn{1}{c|}{0.847}          & 0.711          \\ \hline
\multicolumn{2}{|c|}{MVDNet\_separated\_2}                            & \multicolumn{1}{c|}{0.867}          & \multicolumn{1}{c|}{0.846}          & 0.708          & \multicolumn{1}{c|}{0.876}          & \multicolumn{1}{c|}{0.856}          & 0.656          & \multicolumn{1}{c|}{\textbf{0.886}} & \multicolumn{1}{c|}{\textbf{0.865}} & \textbf{0.722} & \multicolumn{1}{c|}{0.883}          & \multicolumn{1}{c|}{0.856}          & 0.722          & \multicolumn{1}{c|}{0.874}          & \multicolumn{1}{c|}{0.855}          & 0.718          & \multicolumn{1}{c|}{0.867}          & \multicolumn{1}{c|}{0.854}          & 0.712          \\ \hline
\multicolumn{2}{|c|}{MVDNet\_separated\_3}                            & \multicolumn{1}{c|}{0.861}          & \multicolumn{1}{c|}{0.84}           & 0.689          & \multicolumn{1}{c|}{0.868}          & \multicolumn{1}{c|}{0.848}          & 0.699          & \multicolumn{1}{c|}{0.877}          & \multicolumn{1}{c|}{0.856}          & 0.71           & \multicolumn{1}{c|}{\textbf{0.888}} & \multicolumn{1}{c|}{\textbf{0.868}} & \textbf{0.723} & \multicolumn{1}{c|}{0.88}           & \multicolumn{1}{c|}{0.859}          & 0.714          & \multicolumn{1}{c|}{0.879}          & \multicolumn{1}{c|}{0.859}          & 0.713          \\ \hline
\multicolumn{2}{|c|}{MVDNet\_separated\_4}                            & \multicolumn{1}{c|}{0.841}          & \multicolumn{1}{c|}{0.82}           & 0.682          & \multicolumn{1}{c|}{0.848}          & \multicolumn{1}{c|}{0.828}          & 0.694          & \multicolumn{1}{c|}{0.858}          & \multicolumn{1}{c|}{0.838}          & 0.703          & \multicolumn{1}{c|}{0.878}          & \multicolumn{1}{c|}{0.859}          & 0.725          & \multicolumn{1}{c|}{\textbf{0.887}} & \multicolumn{1}{c|}{\textbf{0.868}} & \textbf{0.725} & \multicolumn{1}{c|}{0.879}          & \multicolumn{1}{c|}{0.86}           & 0.723          \\ \hline
\multicolumn{2}{|c|}{MVDNet\_separated\_5}                            & \multicolumn{1}{c|}{0.869}          & \multicolumn{1}{c|}{0.848}          & 0.702          & \multicolumn{1}{c|}{0.867}          & \multicolumn{1}{c|}{0.847}          & 0.699          & \multicolumn{1}{c|}{0.875}          & \multicolumn{1}{c|}{0.855}          & 0.701          & \multicolumn{1}{c|}{0.885}          & \multicolumn{1}{c|}{0.865}          & 0.711          & \multicolumn{1}{c|}{0.887}          & \multicolumn{1}{c|}{0.867}          & 0.711          & \multicolumn{1}{c|}{\textbf{0.887}} & \multicolumn{1}{c|}{\textbf{0.867}} & \textbf{0.712} \\ \hline
\end{tabular} 
}

\vspace{0.2in}
\centering
\caption{Overall Performance of The Mixed Training MVDNet Model: AP of Oriented Bounding Boxes in BEV.}
\label{trainMixed}
\scalebox{0.82}{
\renewcommand{\arraystretch}{1.2} 
\begin{tabular}{|cc|ccc|ccc|ccc|ccc|ccc|ccc|}
\hline
\multicolumn{1}{|c|}{\multirow{2}{*}{Model}} & \emph{offset} & \multicolumn{3}{c|}{0}                                          & \multicolumn{3}{c|}{1}                                          & \multicolumn{3}{c|}{2}                                          & \multicolumn{3}{c|}{3}                                          & \multicolumn{3}{c|}{4}                                          & \multicolumn{3}{c|}{5}                                          \\ \cline{2-20} 
\multicolumn{1}{|c|}{}                       & IoU         & \multicolumn{1}{c|}{0.5}   & \multicolumn{1}{c|}{0.65}  & 0.8   & \multicolumn{1}{c|}{0.5}   & \multicolumn{1}{c|}{0.65}  & 0.8   & \multicolumn{1}{c|}{0.5}   & \multicolumn{1}{c|}{0.65}  & 0.8   & \multicolumn{1}{c|}{0.5}   & \multicolumn{1}{c|}{0.65}  & 0.8   & \multicolumn{1}{c|}{0.5}   & \multicolumn{1}{c|}{0.65}  & 0.8   & \multicolumn{1}{c|}{0.5}   & \multicolumn{1}{c|}{0.65}  & 0.8   \\ \hline
\multicolumn{2}{|c|}{MVDNet\_mixed}                        & \multicolumn{1}{c|}{0.897} & \multicolumn{1}{c|}{0.877} & 0.735 & \multicolumn{1}{c|}{0.896} & \multicolumn{1}{c|}{0.876} & 0.734 & \multicolumn{1}{c|}{0.896} & \multicolumn{1}{c|}{0.877} & 0.734 & \multicolumn{1}{c|}{0.895} & \multicolumn{1}{c|}{0.875} & 0.733 & \multicolumn{1}{c|}{0.895} & \multicolumn{1}{c|}{0.876} & 0.727 & \multicolumn{1}{c|}{0.895} & \multicolumn{1}{c|}{0.875} & 0.726 \\ \hline
\end{tabular}
}
\vspace{-0.1in}
\end{table*}

% mixed training 
\textbf{Mixed training strategy.} Another strategy is to mix an equal amount of paired Radar/Lidar frames with various \emph{offset} together. During training, we use the newly mixed training set to train a MVDNet\_\emph{mixed} model from scratch. The goal is to obtain a MVDNet\_\emph{mixed} model capable of handling Radar/Lidar frames with different \emph{offset}.
The evaluation results are summarized in Table \ref{trainMixed}. We observe that the MVDNet\_\emph{mixed} model achieves similar and good accuracy performance with different IoU settings, indicating its ability to handle Radar/Lidar frames with different \emph{offset}.

Comparing the results of the two strategies, we observe two phenomena. 
Firstly, for aligned Radar/Lidar frames (\emph{offset}=0), the separately fine-tuned MVDNet\_separate\_\emph{0} model achieves slightly better accuracy than the MVDNet\_\emph{mixed} model. 
Secondly, for other unaligned Radar/Lidar frames (\emph{offset}$\in$\{1,2,3,4,5\}), the MVDNet\_\emph{mixed} model achieves better accuracy than separately fine-tuned MVDNet\_separated\_\emph{1/2/3/4/5} models.
% Therefore, the design of a unified MVDNet model capable of handling diverse Radar/Lidar frames involves a trade-off.
Therefore, designing a unified MVDNet model capable of handling diverse Radar/Lidar frames involves a trade-off.
%\vspace{-1em}
\vspace{-0.08in}
\begin{figure}[htbp]
\centerline{\includegraphics[scale=0.5]{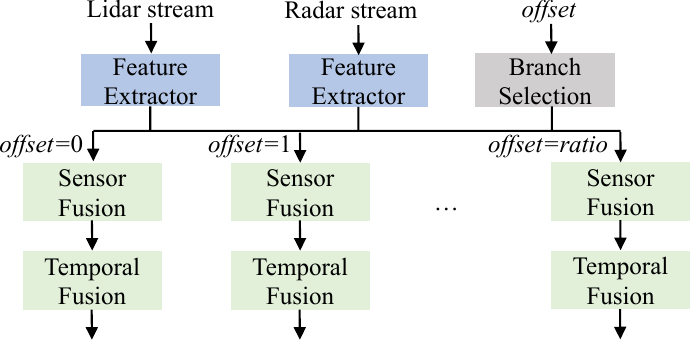}}
\vspace{-0.08in}
\caption{Structure of the multi-branch unified MVDNet.}
\label{branch}
\vspace{-0.1in}
\end{figure}

Based on the separated training strategy, we propose a multi-branch unified MVDNet architecture, as shown in Fig.~\ref{branch}. 
The architecture consists of three main modules. 
The first module consists of feature extractors designed for Radar/Lidar frames. As neural networks tend to extract low-level features in the lower layers \cite{sogaard2016deep}, we leverage this property and utilize the low-level feature extractors as shared feature extractors for Radar/Lidar frames with different \emph{offset}.
The second module is the fusion module, which incorporates multiple branches. Each branch is dedicated to fusing Radar/Lidar frames with a specific \emph{offset}.
The third module introduces a new input \emph{offset} to determine which branch should be chosen for inference.

\subsection{Impact of Historical Information}
\label{ss:history}

%table: align historical frames
\begin{table*}[t]
\centering
\caption{Overall performance of frame skipping method: AP of oriented bounding boxes in BEV.}
\label{skip}
\scalebox{0.811}{
\renewcommand{\arraystretch}{1.2} 
\begin{tabular}{|cc|ccc|ccc|ccc|ccc|ccc|ccc|}
\hline
\multicolumn{1}{|c|}{}                                              & \emph{offset}                 & \multicolumn{3}{c|}{0}                                                                                                               & \multicolumn{3}{c|}{1}                                          & \multicolumn{3}{c|}{2}                                          & \multicolumn{3}{c|}{3}                                          & \multicolumn{3}{c|}{4}                                          & \multicolumn{3}{c|}{5}                                          \\ \cline{2-20} 
\multicolumn{1}{|c|}{\multirow{-2}{*}{Strategy}}                    & IoU                          & \multicolumn{1}{c|}{0.5}                          & \multicolumn{1}{c|}{0.65}                         & 0.8                          & \multicolumn{1}{c|}{0.5}   & \multicolumn{1}{c|}{0.65}  & 0.8   & \multicolumn{1}{c|}{0.5}   & \multicolumn{1}{c|}{0.65}  & 0.8   & \multicolumn{1}{c|}{0.5}   & \multicolumn{1}{c|}{0.65}  & 0.8   & \multicolumn{1}{c|}{0.5}   & \multicolumn{1}{c|}{0.65}  & 0.8   & \multicolumn{1}{c|}{0.5}   & \multicolumn{1}{c|}{0.65}  & 0.8   \\ \hline
\multicolumn{2}{|c|}{\begin{tabular}[c]{@{}c@{}}Without frame skipping,\\ num\_history=4\end{tabular}}  & \multicolumn{1}{c|}{{\color[HTML]{333333} 0.897}} & \multicolumn{1}{c|}{{\color[HTML]{333333} 0.877}} & {\color[HTML]{333333} 0.735} & \multicolumn{1}{c|}{0.898} & \multicolumn{1}{c|}{0.869} & 0.722 & \multicolumn{1}{c|}{0.899} & \multicolumn{1}{c|}{0.877} & 0.724 & \multicolumn{1}{c|}{0.89}  & \multicolumn{1}{c|}{0.867} & 0.722 & \multicolumn{1}{c|}{0.888} & \multicolumn{1}{c|}{0.86}  & 0.709 & \multicolumn{1}{c|}{0.885} & \multicolumn{1}{c|}{0.865} & 0.707 \\ \hline
\multicolumn{2}{|c|}{\begin{tabular}[c]{@{}c@{}}Without frame skipping, \\ num\_history=2\end{tabular}} & \multicolumn{1}{c|}{0.871}                        & \multicolumn{1}{c|}{0.85}                         & 0.725                        & \multicolumn{1}{c|}{0.883} & \multicolumn{1}{c|}{0.863} & 0.72  & \multicolumn{1}{c|}{0.885} & \multicolumn{1}{c|}{0.866} & 0.721 & \multicolumn{1}{c|}{0.885} & \multicolumn{1}{c|}{0.864} & 0.718 & \multicolumn{1}{c|}{0.869} & \multicolumn{1}{c|}{0.856} & 0.712 & \multicolumn{1}{c|}{0.878} & \multicolumn{1}{c|}{0.866} & 0.723 \\ \hline
\multicolumn{2}{|c|}{\begin{tabular}[c]{@{}c@{}}With frame skipping, \\ num\_history=2\end{tabular}}    & \multicolumn{1}{c|}{0.895}                        & \multicolumn{1}{c|}{0.877}                        & 0.724                        & \multicolumn{1}{c|}{0.885} & \multicolumn{1}{c|}{0.865} & 0.715 & \multicolumn{1}{c|}{0.873} & \multicolumn{1}{c|}{0.852} & 0.696 & \multicolumn{1}{c|}{0.887} & \multicolumn{1}{c|}{0.856} & 0.721 & \multicolumn{1}{c|}{0.887} & \multicolumn{1}{c|}{0.857} & 0.7   & \multicolumn{1}{c|}{0.885} & \multicolumn{1}{c|}{0.864} & 0.706 \\ \hline
\end{tabular}
}

\vspace{0.2in}

% Second table
\caption{Overall performance of historical frame alignment strategy: AP of oriented bounding boxes in BEV.}
%Bold numbers represent the best score among all models on Radar/Lidar frames.}
\label{align}
\scalebox{0.82}{
\renewcommand{\arraystretch}{1.2} 
\begin{tabular}{|cc|ccc|ccc|ccc|ccc|ccc|}
\hline
\multicolumn{1}{|c|}{}                           & \textit{offset} & \multicolumn{3}{c|}{1}                                                                                                                                          & \multicolumn{3}{c|}{2}                                                                                                                                          & \multicolumn{3}{c|}{3}                                                                                                                                         & \multicolumn{3}{c|}{4}                                                                                                                                         & \multicolumn{3}{c|}{5}                                                                                                                                          \\ \cline{2-17} 
\multicolumn{1}{|c|}{\multirow{-2}{*}{Strategy}} & IoU             & \multicolumn{1}{c|}{{\color[HTML]{333333} 0.5}}            & \multicolumn{1}{c|}{{\color[HTML]{333333} 0.65}}           & {\color[HTML]{333333} 0.8}            & \multicolumn{1}{c|}{{\color[HTML]{333333} 0.5}}            & \multicolumn{1}{c|}{{\color[HTML]{333333} 0.65}}           & {\color[HTML]{333333} 0.8}            & \multicolumn{1}{c|}{{\color[HTML]{333333} 0.5}}           & \multicolumn{1}{c|}{{\color[HTML]{333333} 0.65}}           & {\color[HTML]{333333} 0.8}            & \multicolumn{1}{c|}{{\color[HTML]{333333} 0.5}}            & \multicolumn{1}{c|}{{\color[HTML]{333333} 0.65}}          & {\color[HTML]{333333} 0.8}            & \multicolumn{1}{c|}{{\color[HTML]{333333} 0.5}}            & \multicolumn{1}{c|}{{\color[HTML]{333333} 0.65}}           & {\color[HTML]{333333} 0.8}            \\ \hline
\multicolumn{2}{|c|}{unaligned}                                    & \multicolumn{1}{c|}{{\color[HTML]{333333} \textbf{0.898}}} & \multicolumn{1}{c|}{{\color[HTML]{333333} \textbf{0.869}}} & {\color[HTML]{333333} \textbf{0.722}} & \multicolumn{1}{c|}{{\color[HTML]{333333} \textbf{0.899}}} & \multicolumn{1}{c|}{{\color[HTML]{333333} \textbf{0.877}}} & {\color[HTML]{333333} \textbf{0.724}} & \multicolumn{1}{c|}{{\color[HTML]{333333} \textbf{0.89}}} & \multicolumn{1}{c|}{{\color[HTML]{333333} \textbf{0.867}}} & {\color[HTML]{333333} \textbf{0.722}} & \multicolumn{1}{c|}{{\color[HTML]{333333} \textbf{0.888}}} & \multicolumn{1}{c|}{{\color[HTML]{333333} \textbf{0.86}}} & {\color[HTML]{333333} \textbf{0.709}} & \multicolumn{1}{c|}{{\color[HTML]{333333} \textbf{0.885}}} & \multicolumn{1}{c|}{{\color[HTML]{333333} \textbf{0.865}}} & {\color[HTML]{333333} \textbf{0.707}} \\ \hline
\multicolumn{2}{|c|}{aligned}                                      & \multicolumn{1}{c|}{{\color[HTML]{333333} 0.889}}          & \multicolumn{1}{c|}{{\color[HTML]{333333} 0.859}}          & {\color[HTML]{333333} 0.694}          & \multicolumn{1}{c|}{{\color[HTML]{333333} 0.895}}          & \multicolumn{1}{c|}{{\color[HTML]{333333} 0.874}}          & {\color[HTML]{333333} 0.732}          & \multicolumn{1}{c|}{{\color[HTML]{333333} 0.88}}          & \multicolumn{1}{c|}{{\color[HTML]{333333} 0.859}}          & {\color[HTML]{333333} 0.718}          & \multicolumn{1}{c|}{{\color[HTML]{333333} 0.879}}          & \multicolumn{1}{c|}{{\color[HTML]{333333} 0.859}}         & {\color[HTML]{333333} 0.705}          & \multicolumn{1}{c|}{{\color[HTML]{333333} 0.856}}          & \multicolumn{1}{c|}{{\color[HTML]{333333} 0.836}}          & {\color[HTML]{333333} 0.687}          \\ \hline
\end{tabular}
}
\vspace{-0.1in}
\end{table*}

In order to navigate the trade-off between latency and accuracy performance, we exploit two strategies for selecting and matching historical frames, aiming to reduce inference latency.

\begin{figure}[htbp]
\centering
\subfigure[The basic method, \emph{num\_history}=2.]{
\begin{minipage}[t]{0.95\linewidth}
\centerline{\includegraphics[scale=0.5]{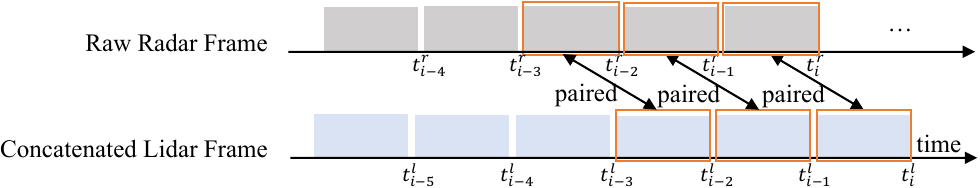}}
\label{hMVDNet}
\end{minipage}
}
%\vspace{-0.5em}
\subfigure[Our frame skipping method, \emph{num\_history}=2.]{
\begin{minipage}[t]{0.95\linewidth}
\centerline{\includegraphics[scale=0.5]{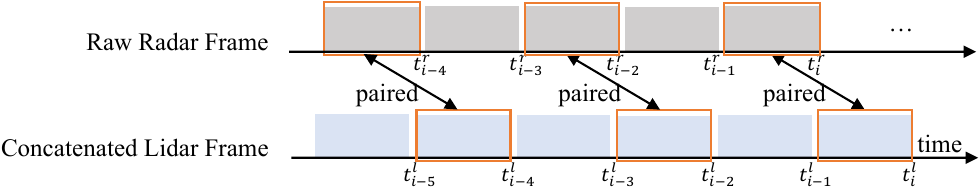}}
\label{hOur}
\end{minipage}
}
\vspace{-0.08in}
\caption{Historical frame selection strategy.}
\vspace{-0.2in}
\end{figure}

\textbf{Historical frame skipping strategy.} The historical frame skipping strategy aims to reduce inference latency while maintaining acceptable accuracy. 
In the basic fusion method (Fig.~\ref{example}), the current and historical paired Radar/Lidar frames are selected consecutively in time. For instance, with a \emph{num\_history}=2 configuration (Fig. 6(a)), the current frame and two consecutive paired frames are chosen as input for one fusion. 
Recognizing the presence of redundant information across consecutive frames, we have explored a frame skipping method that allows us to represent the same amount of information using fewer frames compared to the basic fusion method.
% Recognizing the presence of redundant information between successive frames, we explore a frame skipping method that can use fewer frames to represent more amount of information.
Fig.~\ref{hOur} illustrates our frame skipping method with the same \emph{num\_history}=2 configuration. In this approach, historical paired Radar/Lidar frames are selected every two intervals, effectively skipping several frames. Even though the fusion process only utilize two historical frames, the overall amount of information represented remains equivalent to that of the \emph{num\_history}=4 configuration.
% figure: skipping strategy
\vspace{-0.08in}

Table \ref{skip} shows the evaluation results. 
When considering \emph{num\_history}=2 configuration and a specific \emph{offset} (\emph{offset}$\in$\{0,1,2,...,5\}), our frame skipping method achieves comparable or better accuracy compared to the method without frame skipping, indicating the effectiveness of redundant data.
Additionally, compared to the \emph{num\_history}=4 without skipping configuration, our method \emph{num\_history}=2 with skipping configuration only experiences a slight decrease in accuracy.
The inference latency increases as the number of input frames used for fusion increases. 
Therefore, our method effectively reduces latency with minimal sacrifice in accuracy, offering a novel approach for achieving an accuracy-latency trade-off.
% Table \ref{skip} shows the evaluation results. For Radar/Lidar frames with a specific \emph{offset} (\emph{offset}$\in$\{0,1,2,...,5\}), our frame skipping method achieves comparable or better accuracy compared to the method without frame skipping, using the same \emph{num\_history}=2 configuration.
% Additionally, compared to the \emph{num\_history}=4 configuration, our method with \emph{num\_history}=2 configuration only experiences a slight decrease in accuracy.
% The inference latency increases as the number of input frames used for fusion increases. Therefore, our method effectively reduces latency with minimal sacrifice in accuracy. We expect our strategy can provide a novel approach to achieve a trade-off between accuracy and latency.
%figure: alignment strategy
\vspace{-0.1in}
\begin{figure}[h]
\centerline{\includegraphics[scale=0.5]{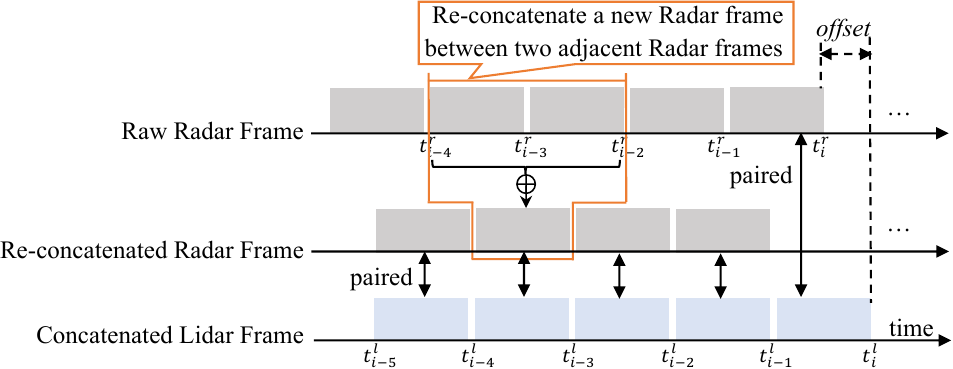}}
\vspace{-0.08in}
\caption{Historical frame alignment strategy, \emph{num\_history}=4.}
\label{hAlign}
\vspace{-0.1in}
\end{figure}

\textbf{Historical frame alignment strategy.} 
In the basic fusion method (Fig.~\ref{example}), misalignment between the current paired Radar/Lidar frames results in misalignment for all other historical paired frames with the same \offset.
% In the basic fusion method (Fig.~\ref{example}), for each 1+{\emph{num\_history}} paired Radar/Lidar frames, if the current paired Radar/Lidar frames are misaligned, all other historical paired frames are also misaligned and exist the same \offset. 
However, we observe that apart from the current frames, other historical concatenated Lidar frames can find their aligned Radar frames.
As shown in Fig.~\ref{hAlign}, while the current concatenated Lidar frame at $t_i^l$ can only be paired with the latest available raw Radar frame arriving at $t_i^r$, each historical concatenated Lidar frame can be aligned perfectly with a reconstructed Radar frame. This new reconstructed Radar frame is obtained by concatenating two adjacent original raw Radar frames.
Based on this observation, we propose a historical frame alignment strategy to align all historical paired Radar/Lidar frames in time. The basic idea is, when selecting each historical paired frame, reconstructing a new Radar frame that aligns with the concatenated Lidar frame according to the Lidar timestamp.

Table \ref{align} compares our strategy with the basic method. We observe that our alignment strategy does not yield any accuracy improvement for temporally unaligned Radar/Lidar frames with a specific \emph{offset}, where \emph{offset}$\in$\{1,2,...,5\}.

\section{Conclusion}
In this paper, we propose techniques to fuse surround Radar/Lidar data by leveraging the faster sampling rate of Lidar instead of the slower Radar. Our approach is based on the widely used MVDNet model for Radar/Lidar fusion-based object detection. 
We aim to maximize the benefits of Lidar's fast sampling and achieve real-time fusion of temporally unaligned Radar/Lidar frames to provide timely road condition information. 
Experimental results demonstrate that our fusion method significantly increases the fusion frequency of temporally unaligned Radar/Lidar data with minimal impact on object detection accuracy.

% \section*{Acknowledgment}
% % We sincerely thank the anonymous reviewers for their valuable feedback. 
% This work is partially supported by the Research Grants Council of Hong Kong (GRF 11208522, 15206221).

\bibliography{reference}
\bibliographystyle{IEEEtran}

\vspace{12pt}

\end{document}